\documentclass[11pt,authoryear]{elsarticle}

\usepackage{amssymb}

\usepackage{graphicx}%
\usepackage{amssymb,amsmath,amsthm}

\newtheorem{theorem}{Theorem}

\usepackage{multirow}%
\usepackage{amsmath,amssymb,amsfonts}%
\usepackage{amsthm}%
\usepackage{mathrsfs}%
\usepackage[title]{appendix}%
\usepackage{hyperref}
\usepackage{manyfoot}%
\usepackage{booktabs}%
\usepackage{algpseudocode}%
\usepackage{listings}%
\usepackage[utf8]{inputenc} 
\usepackage[T1]{fontenc}    
\usepackage{hyperref}       
\usepackage{url}            
\usepackage{booktabs}       
\usepackage{amsfonts}       
\usepackage{nicefrac}       
\usepackage{rotating}
\usepackage{savesym}
\usepackage{graphicx}
\usepackage{adjustbox}
\usepackage[pdftex]{color}
\savesymbol{AND}
\usepackage{epstopdf}
\usepackage{adjustbox}
\usepackage{lscape}

\DeclareMathOperator*{\argmin}{arg\,min}

\usepackage{mathtools}
\usepackage{longtable,pdflscape,booktabs}
\DeclarePairedDelimiterX{\norm}[1]{\lVert}{\rVert}{#1}
\DeclarePairedDelimiterX{\inp}[2]{\langle}{\rangle}{#1, #2}

\usepackage{cleveref}
\usepackage{mathtools}

\usepackage{array,multirow}
\usepackage{graphicx}
\usepackage{subcaption}
\usepackage{setspace}

\makeatletter
\def\ps@pprintTitle{%
  \let\@oddhead\@empty
  \let\@evenhead\@empty
  \let\@oddfoot\@empty
  \let\@evenfoot\@oddfoot
}
\makeatother

\begin{document}

\begin{frontmatter}





%
\title{A Two-Stage Feature Selection Approach for Robust Evaluation
of Treatment Effects in High-Dimensional Observational Data}

\author[mymainaddress]{Md Saiful Islam}

\author[mymainaddress]{Sahil Shikalgar}

\author[mymainaddress]{Md. Noor-E-Alam\corref{mycorrespondingauthor}}
\cortext[mycorrespondingauthor]{Corresponding author}
\ead{mnalam@neu.edu}

\address[mymainaddress]{Department of Mechanical and Industrial Engineering, Northeastern University, Boston, MA 02115, USA}

\begin{abstract}
A Randomized Control Trial (RCT) is considered as the gold standard for evaluating the effect of any intervention or treatment. However, its feasibility is often hindered by ethical, economical, and legal considerations, making observational data a valuable alternative for drawing causal conclusions. Nevertheless, healthcare observational data presents a difficult challenge due to its high dimensionality, requiring careful consideration to ensure unbiased, reliable, and robust causal inferences. To overcome this challenge, in this study, we propose a novel two-stage feature selection technique called, Outcome Adaptive Elastic Net (OAENet), explicitly designed for making robust causal inference decisions using matching techniques. OAENet offers several key advantages over existing methods: superior performance on correlated and high-dimensional data compared to the existing methods and the ability to select specific sets of variables (including confounders and variables associated only with the outcome). This ensures robustness and facilitates an unbiased estimate of the causal effect. Numerical experiments on simulated data demonstrate that OAENet significantly outperforms state-of-the-art methods by either producing a higher-quality estimate or a comparable estimate in significantly less time. To illustrate the applicability of OAENet, we employ large-scale US healthcare data to estimate the effect of Opioid Use Disorder (OUD) on suicidal behavior. When compared to competing methods, OAENet closely aligns with existing literature on the relationship between OUD and suicidal behavior. Performance on both simulated and real-world data highlights that OAENet notably enhances the accuracy of estimating treatment effects or evaluating policy decision-making with causal inference.

\end{abstract}



\end{frontmatter}



\section{Introduction}\label{sec1}

In healthcare research, the Randomized Control Trial (RCT) has  been hailed as the gold standard for assessing the efficacy of interventions and treatments \citep{meldrum2000brief}. Randomization in research studies is a powerful tool for exploring cause-effect relationships between interventions and outcomes. By mitigating bias and ensuring the balance of participant characteristics across groups, randomization enhances our ability to attribute differences in outcomes to the studied intervention. However, RCTs come with notable drawbacks. They can be expensive, time-intensive, and intricate to plan, execute and oversee. Moreover, RCTs might confront practical or ethical limitations, hindering their feasibility or relevance in specific situations. These inherent limitations have led researchers and practitioners to explore alternative avenues for generating valuable insights into treatment effects. One such avenue is the utilization of observational data to draw unbiased causal conclusions \citep{stuart2010matching}. In this context, observational data serves as a pragmatic alternative that can help circumvent some of the practical and ethical hurdles associated with RCTs.

A natural approach to observational studies is matching treated and control units to ensure the balance of underlying background covariates using matching algorithms and calculating the treatment effect of the intervention. However, when it comes to high-dimensional data traditional matching suffers from the curse of dimensionality \citep{abadie:2006, roberts2020adjusting}. Exact matching and coarsened exact matching become less effective as the input dimensionality increases, resulting in a dramatic reduction in the number of viable matches and propensity score matching may inadvertently match highly unrelated units together \citep{king2019propensity}. In addition, researchers often take a ``throw in the kitchen sink'' approach and control for all available covariates just to make sure that the estimate is free of confounding bias \citep{shortreed2017outcome}. In support of this practice, in the past, several researchers have claimed that the largest set of observed pre-treatment covariates protects against unobserved confounding. However, recent literature \citep{shortreed2017outcome,brookhart2006variable,vanderweele2011new} show that controlling for all variables may not lead to the expected reduction in bias, even if it includes a sufficient set of confounders. On the other hand, it inflates the variance of the causal effect estimate. To ensure meaningful comparisons between treated and control groups, employing dimensionality reduction techniques is paramount. By reducing the complexity of the data, these methods help in identifying key covariates and patterns that are crucial for achieving a balanced comparison, and therefore also make unbiased and robust causal conclusions.

\textcolor{black}{To address the above challenge, in this paper, we develop a novel variable selection technique that helps us generate efficient estimates of causal quantities from large-scale, high-dimensional observational data. Specifically, we propose an outcome-adaptive elastic net method carefully crafted with adaptive weights to select the variables that reduce bias and increase the efficiency of the causal effect estimate (i.e., the confounder and outcome predictors). At the same time, the outcome-adaptive method ensures the exclusion of variables that increase the variance of the estimate. The proposed method outperforms the state-of-the-art techniques either by producing better causal effect estimates or producing a comparable estimate in significantly faster computation time.}

\section{Variable Selection in Causality: Challenges and State-of-the-art}
In this section, we provide an overview of causal inference methods under the potential outcome framework, common assumptions, variable selection strategies, our objective and relevant literature.

\subsection{Causality under Potential Outcome Framework} \label{s:methods.1}
In this paper, we consider the potential outcome framework \citep{holland1986stat} and matching methods developed under this framework for identifying treatment effects from observational data. In the matching method, an unbiased estimate of causal inference can be achieved if treatment unit $t \in \mathscr{T}$ is exactly matched with a control unit $c \in \mathscr{C}$ in terms of their covariate set $\mathbf{X} \in \mathcal{X}$,  \textcolor{black}{where $|\mathbf{X}| = P$} \citep{rubin:1983}. However, in most of the applications, it is impossible to achieve exact matching \citep{zubizarreta2012using,rubin:1983,nikolaev2013balance,King:2019}. A wide variety of matching methods are employed to make $(t,c)$ pairs (or subset $\mathcal{C}\subset \mathscr{C}$ and $\mathcal{T} \in \mathscr{T}$) as similar as possible \citep{zubizarreta2012using,zubizarreta2015stable,rubin:1983,nikolaev2013balance} in terms of the covariate set $\mathbf{X}$. One of the popular methods (if not the most popular) \citep{pearl2010foundations,stuart2010matching,zubizarreta2012using} is the propensity score matching (PSM) method \citep{rubin:1983,rubin:1985} that estimates each sample's propensity of receiving treatment \textcolor{black}{(i.e., $Pr(A=1|\mathbf{X})$ where $A \in  \{0,1 \}$ is the treatment status of samples)}. \textcolor{black}{This score is then used to} find the $(t,c)$ pairs or subset $\mathcal{T},\mathcal{C}$ by minimizing the differences in their propensity scores. The matching process is repeated and evaluated iteratively until the desired quality of the matches \textcolor{black}{is} achieved.

\begin{figure}[h]
  \includegraphics[width=\linewidth]{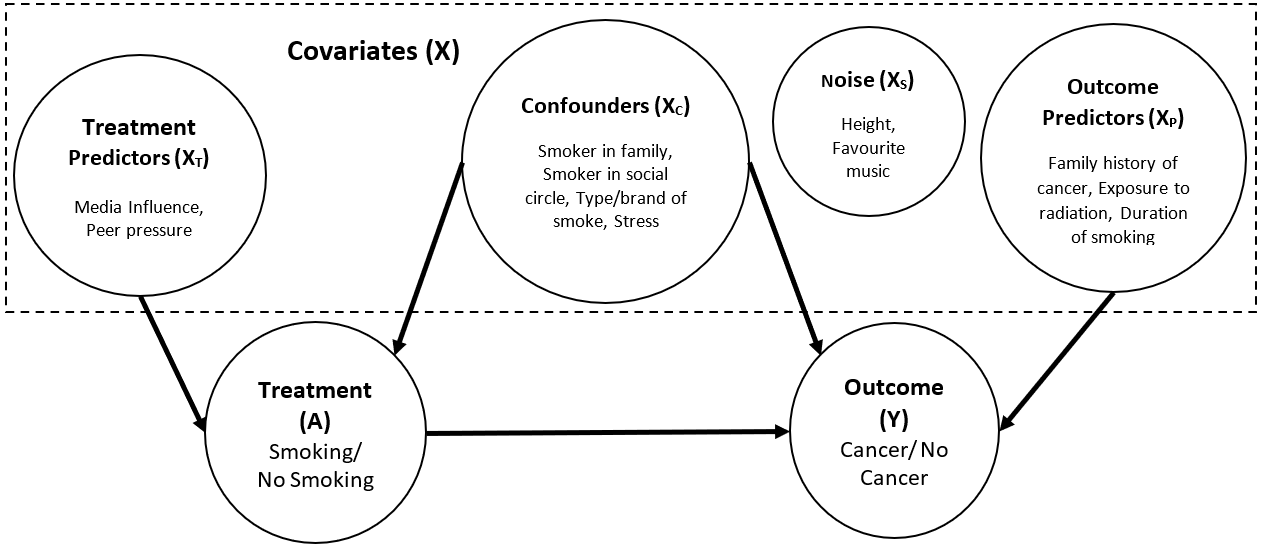}
  \caption{Causal effect of smoking on cancer}
  \label{fig:dummy1}
\end{figure}
\textcolor{black}{For instance, we are interested in estimating the effect of smoking on getting cancer. Due to ethical and legal reasons, we cannot conduct experiments by making people smoke or preventing them from smoking. Therefore, we have to rely on observational data - data that are collected as a natural process. In Figure \ref{fig:dummy1}, we present relations among variables in the study of the cause-effect relation between smoking and cancer. Here, the Treatment ($A$) represents a person's smoking habit ($A$ = 1 if a person smokes and 0 for not smoking) and Outcome ($Y$) is the person's cancer history. The set of covariates ($X$) can be further divided into sub-categories based on their association with treatment and/or outcome. Some variables (i.e., Peer pressure to smoke, Media Influence) will only influence a person's smoking habit which we consider as treatment predictors ($X_T$). Similarly, variables like the Family history of cancer or Radiation exposure are only associated with getting cancer (or no cancer). These variables are not related to a person's smoking habits and are denoted as outcome predictors ($X_P$). In high dimensional data we often have a large number of spurious variables that are neither associated with treatment nor the outcome. We refer to those noise variables as $\mathbf{X}_S$. On the other hand, a subset of variables such as Smokers in the family, and social circle can influence someone's smoking habit as well as having cancer due to secondary smoking. Such variables are called confounders ($X_C$). Now, we can express the treatment allocation mechanism as $A \sim \alpha_C X_C + \alpha_T X_T$ and the outcome model as $Y \sim \beta_C X_C + \beta_P X_P + \textrm{Treatment effect} \times A$. It is important to note that, confounders ($X_C$) appear in both the treatment and outcome model.
}

\textcolor{black}{Propensity score matching estimates a sample's probability of receiving treatment $Pr (A=1|X)$ (i.e., propensity score) and creates matched pairs or subsets between treated and control samples based on the propensity scores. Inclusion of all confounding variables ($X_C$) in the propensity score model or in the matching mechanism ensures an unbiased estimate of the causal quantity considering all such confounders are observed.}

\subsection{Challenges in Variable Selection for Causal Inference} \label{s:methods.2}

While taking the parametric approach, causal effect estimation methods mainly consider two statistical models: outcome model and treatment allocation mechanism, both as functions of pre-treatment variables \citep{ertefaie2018penalization}. Variables that appear in both models are regarded as confounders. The inclusion of such variables in the propensity score model or in the matching mechanism ensures an unbiased estimate of the causal quantity considering all such confounders are observed. \textcolor{black}{Avoiding any confounders will produce a biased estimate.} Ideally, a causal variable selection process must start with the domain knowledge of the underlying treatment allocation mechanism and the predictors of the outcome. Unfortunately, researchers often do not have access to expert knowledge, instead, are burdened with a large set of pre-treatment variables. When domain knowledge is available, the expert selection can result in a significantly large set of variables and particularly, include pure predictors of treatment and pure predictors of outcome. Moreover, high-dimensional observational data can become challenging to comprehend for an expert and may contain spurious variables that do not have any relation with treatment or outcome.

Altogether, the variable selection for causal effect estimation poses three major problems. First, the inclusion of pure treatment predictors \textcolor{black}{($X_T$)} and weak confounders that are strongly related to treatment inflate the variance of the estimate, both asymptotically \citep{rotnitzky2010note} and in a finite sample \citep{brookhart2006variable}. \textcolor{black}{These variables do not contribute to bias reduction and act as noise in matching and including such variables in the matching process (with propensity score or any other matching method) will create mismatched treated-control pairs (or subsets) on important confounders. It will increase the correlation between the matching metric (that will be optimized in the matching process) and the treatment without any reduction in bias. In the example in Figure \ref{fig:dummy1}, variable \textit{Stress} is a weak confounder as it is strongly related to treatment and weakly associated with outcome.
Adding \textit{Stress} and treatment predictors \textit{Peer pressure} and \textit{Media influence} in the propensity score model will introduce random mismatch on confounders like \textit{Smokers in social circles} or \textit{Type of smoke} which will artificially increase the variance of the estimated causal effect of smoking on cancer.} 
In addition, the inclusion of such variables can increase bias if the complete set of confounders is not observed \citep{pearl2012class,wooldridge2016should}. In contrast, selecting the predictors of the outcome \textcolor{black}{($X_P$)} can enhance the efficiency of the causal estimate \citep{brookhart2006variable,rotnitzky2010note}. \textcolor{black}{\citet{brookhart2006variable} argues that in any perceived instance of the data, we might observe a statistically insignificant and small relation between outcomes predictors ($X_P$) and treatment predictors ($X_T$). The inclusion of $X_P$ in the matching process reduces nonsystematic bias because of this `chance association' between $X_P$ and $X_T$. The reduction of nonsystematic bias over different realizations of data makes the treatment effect estimate more efficient.  
} Second, selecting pure treatment predictors \textcolor{black}{($X_T$)} increase the possibility of violating the positivity assumption \citep{stuart2010matching,schnitzer2016ctml}. \textcolor{black}{The positivity assumption states that all samples in the data must have a non-zero probability of receiving treatment.} For example, a strong predictor of treatment can have values for which the probability of receiving treatment is almost zero. Selecting that variable can result in estimating zero probability of receiving treatment in the finite sample and lead to positivity violation. \citet{schnitzer2016ctml} call such a violation an artificial positivity violation. Finally, the inability to discern noise from relevant variables influences the bias and variance of the treatment effect estimate.

\subsection{Objective of Variable Selection for Causal Inference} \label{s:methods.1}
The set of covariates $\mathbf{X}$ can be divided into different sets based on their relations (or the lack of relation) with outcome and treatment. There are different statistical objectives for variable selection in causal inference which use different combinations of these variable sets. For instance, a causal inference expert may decide to include the largest possible set of covariates to control for any possible confounding. Another expert may choose to minimize the variance of the estimate and decide to include all the predictors of the outcome which can be done by \textcolor{black}{a} conditional outcome model $E(Y|T,X)$. Alternatively, we can choose to reduce the bias and use a propensity score model to choose the predictors of the treatment. This approach is suboptimal in terms of bias as it may exclude weak predictors of treatment that are strong predictors of outcome). As we mentioned earlier, the inclusion of the predictors of treatment through the propensity score model may exclude moderate to weak confounders having strong association with outcome and artificially inflate the variance of the estimate. On the other hand, considering just the confounders may not achieve minimum variance estimate in finite-sample. Therefore, the ideal objective of variable selection for causal inference should consider all confounders (variables that are associated with both treatment and outcome) and the predictors of outcome.

\subsection{Relevant Literature} \label{s:methods.2}
To overcome these challenges, the matching mechanism or the Propensity Score (PS) model should include the confounders (predictors of both outcome and treatment) and pure outcome predictors while eliminating pure treatment predictors and noise. There is a vast literature in machine learning that discusses variable selection for prediction purposes, however, there is little work on identifying variables with the objective discussed above.
Bayesian adjustment for confounding (BACR) \citep{wang2012bayesian} is one of the earliest works in variable selection for causal inference. BACR relies on tuning a parameter representing the odds of a variable that is associated with both outcome and treatment. \textcolor{black}{\citet{talbot2015bayesian} shows that the parameter tuning in BACR is very challenging due to its dependence on both the sample size and the data generation process.} \citet{wilson2014confounder} selects the confounders using a decision-theoretic approach. First, they select a group of candidate models by utilizing the posterior credible region of parameters that are identified by a fitted Bayesian Regression model. The final model is determined from the candidate set through the penalization of models that do not incorporate confounders. However, several simulation studies \citep{stijn2012DiscussionBAC,lin2015regularization} reveal that this approach often includes variables that are only associated with treatment resulting in an inflated variance of the causal effect. \textcolor{black}{This approach does not specifically consider the outcome predictors which help reduce the variance of the estimate.} Motivated by the graphical causal inference framework \citep{pearl2009causality} and Bayesian model averaging, \citet{talbot2015bayesian} and \citet{talbot2020generalized} proposed a Bayesian causal effect estimation (BCEE) method. In BCEE, posterior probabilities of each model are considered as weights in the averaging process while probabilities are identified from the conditional outcome model, given treatment, and all observed covariates. However, this approach may \textcolor{black}{leave weak confounders in the model \citep{ertefaie2018penalization} which may reduce an insignificant amount of bias but increase the variance of the estimate}. 

Over the years, several researchers proposed machine learning methods to fit either the treatment model \citep{lee2010improving,westreich2010propensity,austin2013using} or the outcome model \citep{wang2021flame}. \citet{brookhart2006variable} and \citet{schnitzer2016ctml} show that such approaches primarily select variables that are strongly related to either treatment or outcome. \textcolor{black}{If data has moderate to high correlation, these methods have a tendency to select all the variables in the dataset. \citet{koch2020variable} proposed a bi-level approach that combines the outcome and treatment models with the grouped lasso. This approach results in a high inclusion probability on the weak confounders that inflate the variance of the estimate.} Modified penalized regression methods are also utilized in variable selection for causal inference. For instance, \citet{ertefaie2018penalization} used a weighted sum of treatment and outcome model. In addition, they introduced a customized $l_{1}$ penalty function to heavily penalize the treatment predictors. On the other hand, \citet{shortreed2017outcome} uses an adaptive lasso approach which applies less penalty to the outcome predictors. While \citet{ertefaie2018penalization} improves the small sample properties of outcome regularization, the estimated coefficients do not have any causal interpretation, and the parameter selection in \citet{shortreed2017outcome} is designed solely for Inverse Probability of Treatment Weighted (IPTW) estimator. Moreover, both approaches \citep{shortreed2017outcome,ertefaie2018penalization} will perform poorly on correlated data due to the inclusion of $l_1$ penalty and many causal inference applications, especially in healthcare and medicine, the covariates are highly correlated.

In causal inference with the matching method, the variable selection process is followed by a sampling process to match the \textcolor{black}{covariate} distributions of treated and control groups. The matched counterparts of the treated units in the control group are interpreted as counterfactuals, and the average treatment effect on treated (ATT) is estimated by comparing the outcomes of every matched pair \citep{stuart2010matching}. One of the popular matching methods is the Nearest Neighbor Matching (NNM) \citep{rubin:1973a,rosenbaum:2017} which forms a treated-control pair by matching a treated unit to its nearest control unit based on some prespecified distance metrics. Some commonly used NNM matching methods are Propensity Score Matching (PSM) \citep{rubin:1983,rubin:1985}, Coarsened Exact Matching (CEM) \citep{king:2011}, Mahalanobis Distance Matching (MDM) \citep{rubin:1979}, and Genetic Matching (GM) \citep{sekhon2012matching}. In the case of high dimensional data, these widely used NNM methods perform poorly \textcolor{black}{as they do not include any algorithmic mechanism for variable selection. They either heavily rely on domain experts or include all the covariates in the data regardless of their effect on the causal inference}. For instance, PSM projects the \textcolor{black}{covariate set (either identified by domain experts or all covariates in the data)} into one dimension disregarding the neighborhood structure of the dataset, and the matched units in projected dimension often differ on important covariates in actual dimension. MDM is prone to bias in high dimensions since it imposes parametric assumptions, so does the GM. In the presence of a large number of covariates, to obtain a good match, CEM discards a lot of samples. In addition, recent results show that the bias of an NNM estimator increases at a rate of $O(N^{-1/P})$ where $N$ is the number of samples and $P$ is is the number of variables \citep{abadie:2006}. Dimension reduction techniques like Principle Component Analysis (PCA) have been studied extensively in Machine Learning literature where the main objective is to improve the models' predictive accuracy by retaining the information content. 
However, in observational studies using matching methods, the objective is to ensure the local neighborhood structure of the data as a treated sample is essentially matched to a control sample within its locality.

\subsection{Contribution} \label{s:methods.2}

In this paper, we propose a unique modeling framework referred to as Outcome Adaptive Elastic Net (OAENet) specifically designed for causal inference to select the confounders and outcome predictors for inclusion in the propensity score model or in the matching mechanism. OAENet provides two major advantages over \textcolor{black}{many} existing methods: \textcolor{black}{it performs superiorly on correlated and high dimensional data, and it complies with objective of variable selection by including right types of variables}. \textcolor{black}{OAENet can be used as a pre-processing step to improve current approaches in NNM methods by identifying confounding variables whereas current projection-based dimensionality reduction techniques (i.e., PSM and PCA) may inflate variance and bias by including the treatment and outcome predictors, respectively. As the variables selected by OAENet will be used in the matching process, the inclusion of $l_2$ penalty in OAENet will produce stable models for correlated data which will contribute to reduced variability in the estimate.} We discuss the oracle properties of the proposed OAENet. We also compare its performance with state-of-the-art variable selection techniques for causal inference on simulated data. Our preliminary analysis shows that the OAENet \textcolor{black}{achieves either better bias-variance results or has significant computational advantage. In addition, it has} better accuracy in selecting variables compared to the benchmark methods. \textcolor{black}{We} apply the proposed Outcome Adaptive Elastic Net (OAENet) on National Survey of Drug Use and Health (NSDUH) data to identify the effect of opioid use disorder on suicidal behavior.

The remainder of the paper is organized as follows. In section \ref{cause_high_dim}, we discuss causal inference under potential outcome framework, the objective of variable selection for causal inference, and review adaptive elastic-net. In section \ref{out_ad_enet}, we outline the proposed Outcome Adaptive Elastic Net (OAENet) framework. We present the empirical performance of the proposed methods and compare it with alternative approaches on simulated data in section \ref{high_dim_exp}. In section \ref{case}, we provide a case study. The case study evaluates the effect of Opioid Use Disorder(OUD) on suicidal behavior and compares the result of the proposed OAENet to state-of-the-art literature. Finally, we provide the concluding remarks in section \ref{high_dim_conc}.

\section{Causal Inference from High-dimensional Data} \label{cause_high_dim}
\textcolor{black}{In this section, we introduce the adaptive elastic net and present the proposed outcome adaptive elastic net along with its oracle properties.}

\subsection{Adaptive Elastic Net} \label{s:methods.1}
Adaptive elastic-net proposed by \citet{zou2009adaptiveEnet} is an extension of the popular model selection and estimation method of Elastic-net \citep{zou2005Enet}. Elastic-net combines the automatic variable selection property of $l_1$ regularization and the stabilizing property of $l_2$ regularization. Adaptive elastic-net further improves the finite sample performance of elastic-net by introducing an adaptive weight in the $l_1$ penalty. Under weak regularity condition, \citet{zou2009adaptiveEnet} showed that the adaptive elastic-net meets the oracle properties which implies that this adaptive procedure selects the right variables with high probability (i.e., consistency) and estimated coefficients are asymptotically normal. An elastic-net estimator takes the following form:  
\begin{align}
    \hat{\boldsymbol{\beta}}(enet) = (1+\frac{\lambda_2}{n})\left\{ {\argmin}_{\boldsymbol\beta} \left \Vert\mathbf{y}-\mathbf{X\beta}  \right\Vert^2_2 + \lambda_2\left \Vert\boldsymbol{\beta}\right\Vert_2^2 + \lambda_1 \left \Vert\boldsymbol{\beta}\right\Vert_1^1 \right\} \label{enet}
\end{align}
where, $n$ is the sample size, $\mathbf{X}$ \textcolor{black}{is the set of covariates} and $\mathbf{y}$ \textcolor{black}{is the outcome}, and $\lambda_1$, $\lambda_2$ are the regularization penalties. Adaptive elastic-net first computes the elastic-net estimator $\hat{\boldsymbol\beta}(enet)$ as defined in \eqref{enet}, and then uses the estimator as the adaptive weights $\hat{w}_j$ for all \textcolor{black}{$P$} variables as the following. 
\begin{align}
    \hat{w}_j = \left(|\hat{\beta}_j(enet)|\right)^{-\gamma}, \quad \quad j = 1,2,\cdots,P
\end{align}
Here, $\gamma$ is a positive constant. Now, the adaptive weights $\hat{w}_j$ \textcolor{black}{are} multiplied with the $l_1$ penalty to create variable and adaptive penalties among the features. Therefore, the adaptive elastic-net takes the following form:
\begin{align}
    \hat{\boldsymbol{\beta}}(adenet) = (1+\frac{\lambda_2}{n})\left\{ {\argmin}_{\boldsymbol\beta} \left \Vert\mathbf{y}-\mathbf{X\beta}  \right\Vert^2_2 + \lambda_2\left \Vert\boldsymbol{\beta}\right\Vert_2^2 + \lambda_1 \sum_{j=1}^p \hat{w}_j\left \Vert\boldsymbol{\beta}\right\Vert_1^1 \right\} \label{adenet} 
\end{align}
As we can see from equation \eqref{adenet}, if we consider zero $l_2$ penalty then we can recover the adaptive lasso proposed by \citet{zou2006adaptiveLas}.

\subsection{Outcome Adaptive Elastic Net} \label{out_ad_enet}
 
As we mentioned before, the ideal objective for variable selection for causal inference is to select $\mathbf{X}_C$ and $\mathbf{X}_P$, and remove $\mathbf{X}_T$ and $\mathbf{X}_S$. To that end, we modify the adaptive elastic-net to accommodate this objective and propose the Outcome Adaptive Elastic Net (OAENet). In this paper, we confine our focus to the binary treatment. 

Adaptive elastic-net provides the opportunity to create variable penalties for different covariates. We use this to connect the two models: the outcome model and the treatment model. In the first step, we consider the outcome model, however, instead of elastic-net regularization, we use ordinary linear regression to identify the strength of the variables' association to the outcome. 
\textcolor{black}{The purpose of Adaptive Elastic Net is to select desired sets of variables, however, in the first step of OAENet, we do not aim to select variables. Our purpose in
the first step is to estimate the coefficients of variables in predicting outcomes. We will use the
estimated coefficient to create a variable penalty function in step 2.  }
\begin{align}
       \textrm{\textit{Step 1 - Outcome Model:}    } \hat{\boldsymbol{\beta}}(OLS) & = \argmin_{\boldsymbol\beta} \left \Vert\mathbf{y}-\mathbf{X\beta}  \right\Vert^2_2 \\
        & = \argmin_{\boldsymbol\beta} l_n(\boldsymbol{\beta}; y, \mathbf{X}) \textrm{ (MLE of $\beta$) } 
\end{align}

Let us assume, $\hat{\boldsymbol{\beta}}(OLS)$ is the maximum likelihood estimate for the coefficients of the variables when Ordinary Least Squares regression (OLS) is used on outcome $y$: $\hat{\boldsymbol{\beta}}(OLS) = \argmin_{\boldsymbol\beta} l_n(\boldsymbol{\beta}; y, \mathbf{X})$. Now, in the second step, considering binary treatment, we use a logit model with adaptive 
elastic-net regularization as our treatment model. 
\begin{align}
    \textrm{\textit{Step 2 - Treatment Model:}    } \hat{\boldsymbol{\alpha}}_{(OAENet)} = \left( 1+\frac{\lambda_2}{n} \right) \Biggl [ {\argmin}_{\boldsymbol\alpha} \biggl \{ \sum_{i=1}^n  ( -a_i(x_i^T\alpha) +   \mspace{15mu} \notag\\
   log(1+e^{x_i^T\alpha})  ) + \lambda_2\left \Vert\boldsymbol{\alpha}\right\Vert_2^2 + \lambda_1 \sum_{j=1}^p \hat{w}_j \left \Vert\boldsymbol{\alpha}\right\Vert_1^1 \biggr \} \Biggr ] \label{oaenet}
\end{align}
where, $a_i$ is the indicator of treatment status, $\hat{w}_j = (|\hat{\beta}(OLS)_j|)^{-\gamma}$ with $\gamma > 1$. Here, we multiply the inverse of the coefficients of the outcome model with $l_1$ penalty. It \textcolor{black}{applies} smaller penalties to variables that are highly associated with the outcome. Therefore, this model is more likely to \textcolor{black}{include} variables that are confounders and outcome predictors. \textcolor{black}{It is important to note that in OAENet, we do not need to make any additional assumptions than the assumptions in Adaptive Elastic Net. In addition, in step 1, we presented the outcome model considering the continuous outcome, however, we can use a generalized linear model with appropriate link function, without loss of generality, to accommodate any types of outcomes. }

\begin{theorem}
Suppose $\frac{\lambda_1}{\sqrt{n}} \xrightarrow{} 0, \lambda_1 n^{(\gamma-1)/2} \xrightarrow{} \infty, \frac{\lambda_2}{\sqrt{n}} \xrightarrow{} 0,$ $ A=\left\{j, \beta_j \neq 0 | j \in X_P \cup X_C\right\}$, and $\hat{A}(\lambda_1,\lambda_2) = \left\{j:\hat{\beta}_{OAENet}(\lambda_1,\lambda_2) \neq 0 \right\}$. Then, under mild regularity conditions, the proposed OAENet has the oracle Properties by satisfying the following:
  \begin{itemize}
    \item Consistency in variable selection: $\lim_{n\xrightarrow{}\infty}$ Pr $\left(\hat{A}(\lambda_1,\lambda_2) = A\right) = 1$
   
    \item Asymptotic Normality: $\sqrt{n}\left(\hat{\beta}_{OAENet} (\lambda_1,\lambda_2) - \beta^*(\lambda_1,\lambda_2)\right) \sim N(0, C^{-1}_{11}(\lambda_1,\lambda_2))$
\end{itemize}
\end{theorem}

\begin{proof}
Let's assume $C_n = \frac{1}{n}X^TX \rightarrow C$, where $C$ is a positive definite matrix and $|A| = p_0 < P$ where $P$ is the total number of variables in the covariate set. Without loss of generality, we can say that $A = \left \{ 1,2, \cdots, p_0 \right \}$.
Let $C =  \begin{pmatrix}
  C_{11} & C_{12}\\ 
  C_{21} & C_{22}
\end{pmatrix}$
where $C_{11}$ is a $P_0 \times P_0$ matrix. $C_{11}$ is also known as the Fisher Information Matrix.

Now, based on the argument provided by \citet{ghosh2011grouped}, we can modify the data $(y, X)$ into an artificial dataset $(y^*, X^*)$ using the following condition:

For fixed elastic net penalties $(\lambda_1, \lambda_2)$, define $C^*$ as $C^*_n = \frac{1}{n}X^{*T}X^* = C_n + \frac{\lambda_2^{(n)}}{n}$ $\rightarrow C$, where $C$ is a positive definite matrix and $\lambda_2/n \rightarrow 0$ asymptotically.

Let's define $A^*=\left\{j, \hat{\beta_j}^* \neq 0 \right\}$. Now, using the augmented matrix, we can write the adaptive elastic net in equation \eqref{adenet} (with outcome adaptive penalties $\hat{w}_j$ as proposed in equation \eqref{oaenet}) into an Outcome-adaptive lasso problem of \eqref{adalasso}. 
\begin{align}
    \hat{\boldsymbol{\beta}}(adenet) = (1+\frac{\lambda_2}{n})\left\{ {\argmin}_{\boldsymbol\beta} \left \Vert\mathbf{y}^*-\mathbf{X^*\beta}  \right\Vert^2_2 + \lambda_1 \sum_{j=1}^p \hat{w}_j\left \Vert\boldsymbol{\beta}\right\Vert_1^1 \right\} \label{adalasso} 
\end{align}

\citet{shortreed2017outcome} shows that the Outcome-adaptive lasso problem can achieve the oracle properties. Therefore, we can claim that the proposed Outcome Adaptive Elastic Net (OAENet) also achieves the oracle property by satisfying the consistency in variable selection and asymptotic normality. 
\end{proof}
\textcolor{black}{The complete proof is provided in the \textbf{Online Supplement S1 Appendix B.}}

\section{Numerical Experiment} \label{high_dim_exp}
In this section, we discuss the performance of the proposed Outcome Adaptive Elastic Net (OAENet) on simulated data and compare it with state-of-the-art methods.

\subsection{Simulation Design and Scenarios} \label{s:methods.1}

We design the simulation study based on the scenarios designed by \citet{shortreed2017outcome} and \citet{ertefaie2018penalization}. We consider two scenarios where covariates are generated from multivariate Gaussian distribution. We confine our focus to the realm of binary treatment and continuous outcomes. For each scenario, we use two correlation structures: independent covariates (correlation = 0) and strongly correlated (correlation = 0.5). For all the scenarios, we simulate $n=1000$ data points, use $\gamma = 3$ and select the penalty parameters with 5-fold cross-validation \textcolor{black}{at one standard error (i.e., 1 SE). Regarding the penalty parameters ($\lambda_i$), we considered 1 SE as it is commonly used is machine learning literature, however, user can choose parameters with minimum errors. In choosing $\gamma$, we recommend using the guideline provided by \citet{shortreed2017outcome}}. For all the scenarios, data is normalized before using it for variable selection.

\textbf{Scenario 1} is adapted from \citet{shortreed2017outcome}. The data-generating equations are the following:
\begin{itemize}
    \item Covariates: $X_1, X_2, \cdots,X_{100} \stackrel{iid}{\sim} N(0,1)$ with correlation $\rho = 0$ (Scenario 1A) and $\rho = 0.5$ (Scenario 1B)
    \item Treatment: $A \sim Bernoulli( p = Expit(X_1 + \cdots + X_{10} + X_{21}+\cdots+X_{30}))$ where, $Expit (x) = \frac{e^x}{1+e^x} $
    \item Outcome: $y = TE.A + 0.6X_1+0.6X_2+\cdots+0.6X_{20}$ with true treatment effect $TE = 0.5$ and $A=0$ when unit does not receive treatment, $A=1$ when it receives treatment
\end{itemize}

\textbf{Scenario 2} is adapted from \citet{ertefaie2018penalization}. The data-generating equations are the following:
\begin{itemize}
    \item Covariates: $X_1, X_2, \cdots,X_{100} \stackrel{iid}{\sim} N(0, \sigma^2=4)$ with correlation $\rho = 0$ (Scenario 2A) and $\rho = 0.5$ (Scenario 2B)
    \item Treatment: $A \sim Bernoulli( p = Expit(0.5X_1-X_2+0.3X_5-0.3X_6+0.3X_7-0.3X_8))$ where, $Expit (x) = \frac{e^x}{1+e^x} $
    \item Outcome: $y = TE.A + 2X_1+2X_2+5X_3+5X_4$ with true treatment effect $TE = 1.0$ and $A=0$ when unit does not receive treatment, $A=1$ when it receives treatment
\end{itemize}

We compare the performance of the proposed method with the following techniques:
\begin{itemize}
    \item \texttt{BCEE}: Bayesian Causal Effect Estimation method proposed by \citet{talbot2015bayesian} using the \texttt{R} package BCEE \citep{bcee2015package}
    \item Outcome adaptive lasso (\texttt{OLas}): Outcome adaptive lasso proposed by \citet{shortreed2017outcome}
    \item \texttt{BACR}: Bayesian Adjustment for Confounding proposed by \citet{wang2012bayesian} with the \texttt{R} package BACR \citep{bacr2015package}
    \item \texttt{Bor(T)}, \texttt{Bor(Y)}: Identifying the treatment predictors - \texttt{Bor(T)} and the outcome predictors \texttt{Bor(Y)} using random forest with the \texttt{R} package Boruta \citep{boruta2020package} 
\end{itemize}

In addition to the above-mentioned methods, we include the following three benchmarks to evaluate the performance of the proposed technique. As we are simulating the dataset, we take advantage of our knowledge of the true relation between variables, treatment, and outcome.
\begin{itemize}
    \item Target (\texttt{Targ}): Estimating the causal effect with the set of target variables $\mathbf{X}_C$ and $\mathbf{X}_P$. \textcolor{black}{Ideally, a variable selection method should select $\mathbf{X}_C$ and $\mathbf{X}_P$, and produce causal effect estimate close to \texttt{Targ}.}
    \item Confounders (\texttt{Conf}): The set of confounders, variables that are associated with both treatment and outcome ($\mathbf{X}_C$).
    \item Potential confounders (\texttt{Pot.Conf}): Set of potential confounders, variables that are only associated with treatment ($\mathbf{X}_C$ and $\mathbf{X}_T$).
\end{itemize}

\subsection{Discussion} \label{s:methods.2}
We evaluate the performance of the proposed Outcome Adaptive Elastic Net (OAENet) in terms of three metrics: bias and variance of the estimate, the ability to select the target variables\textcolor{black}, {and computational time}. In this experiment, our target is to select the variables that are associated with both treatment and outcome ($\mathbf{X}_C$), and the variables that are associated with only outcome ($\mathbf{X}_P$). For each scenario, we generate 1000 datasets and use the variable selection techniques to select appropriate variables for inclusion in the propensity score model. Using the selected variables, we perform nearest neighbor matching \citep{stuart2010matching} and compute the average treatment effect on treated (ATT) from the matched samples. We also compute the proportion of the time a variable is selected by the variable selection techniques \textcolor{black}{and the time taken by each method to select variables and estimate the ATT}. \textcolor{black}{We used a Dell Precision 7510 workstation with Intel Core i7-6820HQ CPU running at 2.70 GHz, and 32 GB RAM.}

\begin{figure}[ht!]
\centering
    \begin{subfigure}[b]{0.45\textwidth}
    \centering
    \includegraphics[width=0.96\textwidth]{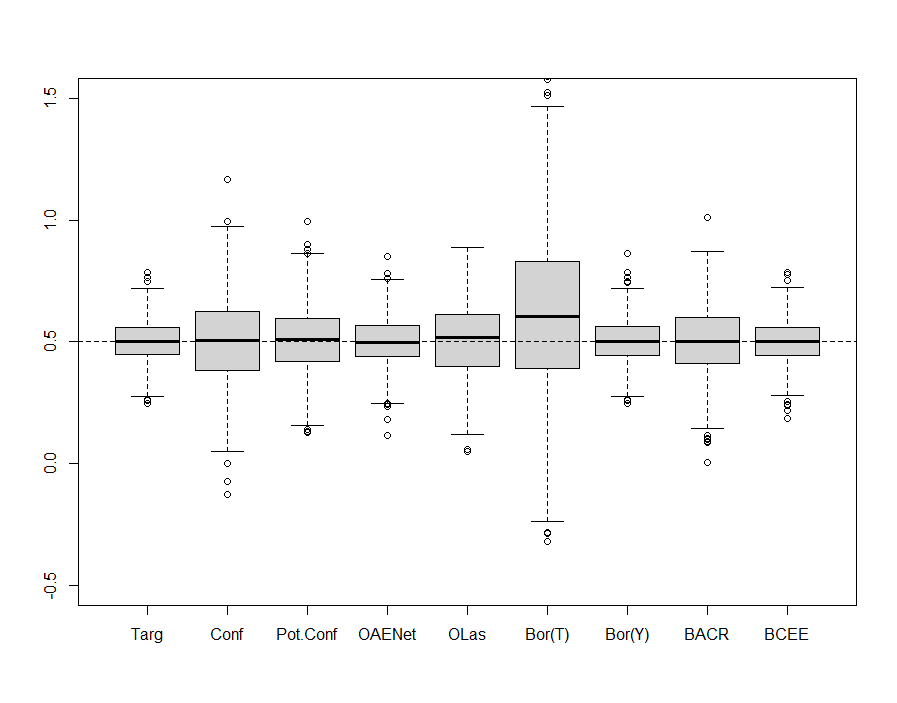}
    \caption{}
    \end{subfigure}
    \begin{subfigure}[b]{0.45\textwidth}
    \centering
    \includegraphics[width=0.94\textwidth]{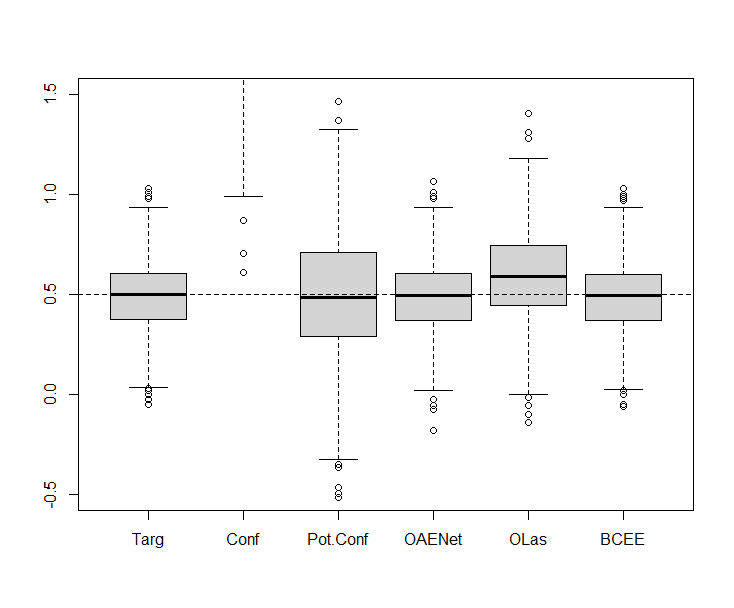}
    \caption{}
    \end{subfigure}  
    \begin{subfigure}[b]{0.45\textwidth}
    \centering
    \includegraphics[width=1\textwidth]{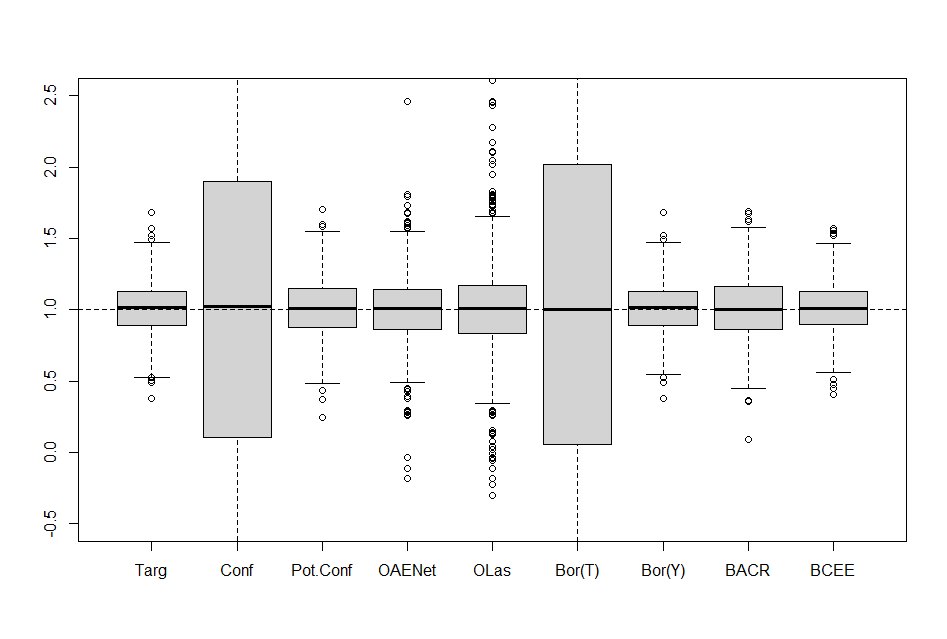}
    \caption{}
    \end{subfigure}
    \begin{subfigure}[b]{0.45\textwidth}
    \centering
    \includegraphics[width=1\textwidth]{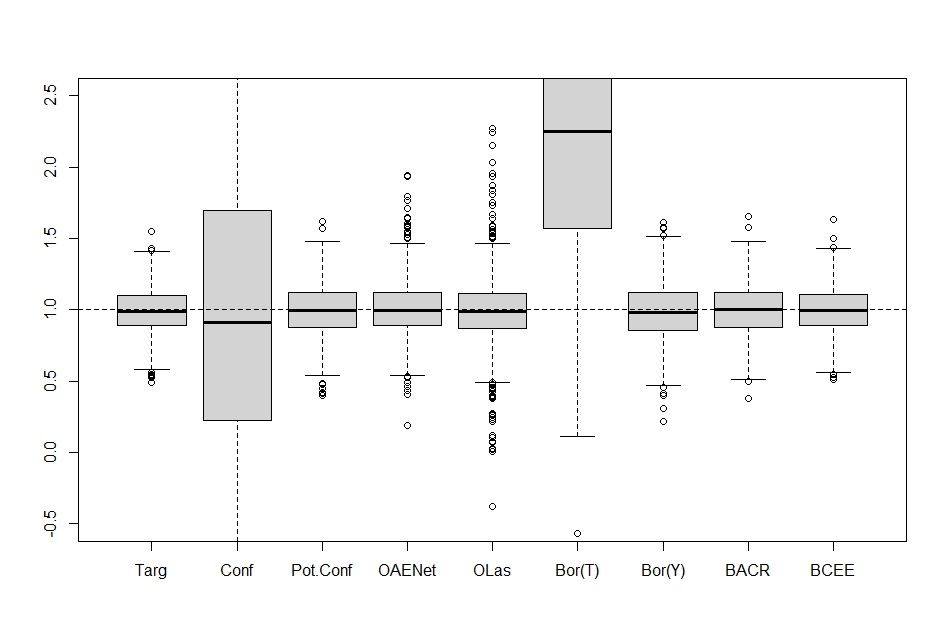}
    \caption{}
    \end{subfigure}
    \caption{Average Treatment effect on the treated (ATT) for (a) Scenario 1A (b) Scenario 1B (c) Scenario 2A (d) Scenario 2B. The horizontal dashed line indicates the true treatment effect.}
    \label{fig:att}
\end{figure}

In Figure \ref{fig:att}, we present the boxplot of the ATT computed for all the scenarios. The objective is to estimate the ATT as close to the target (presented as \texttt{Targ} in Figure \ref{fig:att}). As we can see, overall the proposed Outcome Adaptive Elastic Net (presented as \texttt{OAENet}) and \texttt{BCEE} closely follows the target \textcolor{black}{(i.e., \texttt{Targ}) in all scenarios with \texttt{BCEE} performing marginally better than \texttt{OAENet}.}
\texttt{BACR}, \texttt{Bor(T)}, and \texttt{Bor(Y)} are omitted in the Figure \ref{fig:att}(b) as they failed to calculate the treatment effect. It is important to note that, scenario 1B has 100 variables with 50 noisy variables and a high correlation ($\rho =0.5$). In such situations, \texttt{BACR}, \texttt{Bor(T)}, and \texttt{Bor(Y)} select almost all of the variables, including the spurious variables. It is almost impossible to find a sufficiently large matched set of samples from 1000 data points, therefore, these methods fail to calculate ATT for scenario 1B. However, the proposed outcome adaptive method along with outcome adaptive lasso and \texttt{BCEE} can identify confounders and the outcome predictors, and calculate ATT. On the other hand, for scenarios 2A and 2B in Figures \ref{fig:att} (c) and \ref{fig:att} (d) respectively, \texttt{OAENet} perform similarly to other competing methods. 


\textcolor{black}{A major difference between scenarios 1 and 2 is the number of treatment and outcome predictor variables: 30 vs 6 treatment predictors and 20 vs 4 outcome predictors respectively for scenarios 1 and 2. To have a closer look into the performance of the variable selection methods in those scenarios, we provide the bias and variance of the treatment effect estimates in Table \ref{biasvar}, and the time taken to select variables and estimate ATT in Table \ref{time}. From Table \ref{biasvar}, we can see that when there is no correlation between variables (Scenario 1A and 2A) or a smaller number of predictor variables (Scenario 2A and 2B), \texttt{Bor(Y)} and \texttt{BCEE} perform better than \texttt{OAENet} in terms of bias and variance. For a large number of predictor variables and high correlation scenario - Scenario 1B, we can see that \texttt{OAENet} (variance: 0.0336, bias: 0.0037) outperforms all other methods except \texttt{BCEE} (variance: 0.03063 and bias: 0.0013). At the same time, \texttt{Bor(T)}, \texttt{Bor(Y)}, and \texttt{BACR} fail to estimate the treatment effect in this scenario. Even though \texttt{BCEE} and \texttt{Bor(Y)} perform slightly better than \texttt{OAENet} regarding bias and variance, Table \ref{time} Scenario 1B shows that \texttt{OAENet} (0.8 sec) can complete variable selection along with ATT estimation is more than 200 times faster than \texttt{BCEE} (164.2 sec). \texttt{OAENet} also significantly outperforms \texttt{Bor(Y)} in computation time in all the scenarios. The computational advantage of \texttt{OAENet} over \texttt{BCEE} and \texttt{Bor(Y)} is significant even with 1000 samples and 100 variables considered in the simulated experiment. The use of \texttt{OAENet} will be most beneficial in big data observational studies as it can produce causal effect estimates comparable to the best methods and it is highly scalable whereas other state-of-the-art competing methods become extremely intractable for larger datasets.}

\begin{sidewaystable}
\small
\caption{Bias and Variance of estimated treatment effect by different variable selection methods.}\label{biasvar}
\begin{adjustbox}{scale=0.9,center}
\begin{tabular*}{\textheight}{@{\extracolsep\fill}lcccccc}
\toprule%
\textbf{} & \textbf{OAENet} & \textbf{OLas} & \textbf{Bor(T)} & \textbf{Bor(Y)} & \textbf{BACR} & \textbf{BCEE} \\
\midrule
\textbf{Scenario 1A} & ~ & ~ & ~ & ~ & ~ & ~ \\ \hline
        \textbf{Variance} & 0.0090 & 0.0218 & 0.1103 & 0.0071 & 0.0201 & 0.0073 \\ 
        \textbf{Bias} & 0.0031 & -0.0072 & -0.1198 & -0.0008 & -0.0036 & -0.0007 \\ 
        \textbf{Lower 95\% CI} & -0.1129 & -0.2632  & -0.8893  & -0.0344  & -0.2544  & -0.0887  \\ 
        \textbf{Upper 95\% CI} &  0.1191 &  0.2488 &  0.6497 &  0.0328 &  0.2472 &  0.0873 \\
        \hline
        \textbf{Scenario 1B} & ~ & ~ & ~ & ~ & ~ & ~ \\ \hline
        \textbf{Variance} & 0.0336 & 0.0517 & ~ & ~ & ~ & 0.03063 \\ 
        \textbf{Bias} & 0.0037 & -0.1044 & - & - & - & 0.0013 \\ 
       \textbf{Lower 95\% CI} & -0.1775 & -0.4551  & ~ & ~ & ~ & -0.1766  \\
       \textbf{Upper 95\% CI} &  0.1850 &  0.2463 & ~ & ~ & ~ &  0.1792 \\
       \hline
        \textbf{Scenario 2A} & ~ & ~ & ~ & ~ & ~ & ~ \\ \hline
        \textbf{Variance} & 0.0530 & 0.1114 & 2.0983 & 0.0317 & 0.0441 & 0.0321 \\ 
        \textbf{Bias} & 0.0038 & -0.0045 & -0.0313 & -0.0002 & -0.002 & 0.0004 \\ 
       \textbf{Lower 95\% CI} & -0.2817  & -0.5795  & -2.847  & -0.0836  & -0.3132  & -0.148 \\ 
       \textbf{Upper 95\% CI} &  0.2894 &  0.5704 &   2.7842 &  0.0831 &  0.3092 &  0.1488 \\ 
       \hline
        \textbf{Scenario 2B} & ~ & ~ & ~ & ~ & ~ & ~ \\ \hline
        \textbf{Variance} & 0.0396 & 0.0650 & 1.0806 & 0.0412 & 0.0329 & 0.0271 \\ 
        \textbf{Bias} & -0.01715 & 0.006436 & -1.31223 & 0.002098 & -0.00703 & -0.00112 \\ 
        \textbf{Lower 95\% CI} & -0.2559 & -0.3933 & -3.3346  & -0.298  & -0.2611  & -0.1404 \\ 
        \textbf{Upper 95\% CI} &  0.2216 &  0.4061 & 0.7102&  0.3022 &  0.2471 &  0.1381 \\ \hline

\end{tabular*}
\end{adjustbox}
\end{sidewaystable}

\begin{table}[h]
\caption{Average Computational time (seconds) taken by different variable selection methods to select variables and estimate ATT.}\label{time}
\begin{tabular*}{\textwidth}{@{\extracolsep\fill}lcccccc}
\toprule%
\textbf{} & \textbf{OAENet} & \textbf{OLas} & \textbf{Bor(T)} & \textbf{Bor(Y)} & \textbf{BACR} & \textbf{BCEE} \\
\midrule
        \textbf{Scenario 1A} & 0.627 & 0.136 & 33.318  & 29.476 & 12.576 & 122.523 \\  \hline
        \textbf{Scenario 1B} & 0.815 & 0.205 & - & - & - & 164.278 \\ \hline
        \textbf{Scenario 2A} & 0.355 & 0.122 & 25.881 & 31.889 & 11.347 & 80.573 \\ \hline
        \textbf{Scenario 2B} & 0.425 & 0.147 & 57.477 & 62.409 & 10.858 & 75.395 \\ \hline
\end{tabular*}

\end{table}

In addition, recent literature \citep{brookhart2006variable,shortreed2017outcome,ertefaie2018penalization} suggest that considering only the treatment predictors does not reduce bias but inflates variance of the estimate. Our experiment also shows similar results as we can see in Figures \ref{fig:att} (a), (c), and (d), \texttt{Bor(T)} which considers only the treatment predictors have the most variance in ATT among all the variable selection techniques. 

\begin{figure}[ht!]
    \centering
    \begin{subfigure}[b]{0.45\textwidth}
    \centering
    \includegraphics[width=1\textwidth]{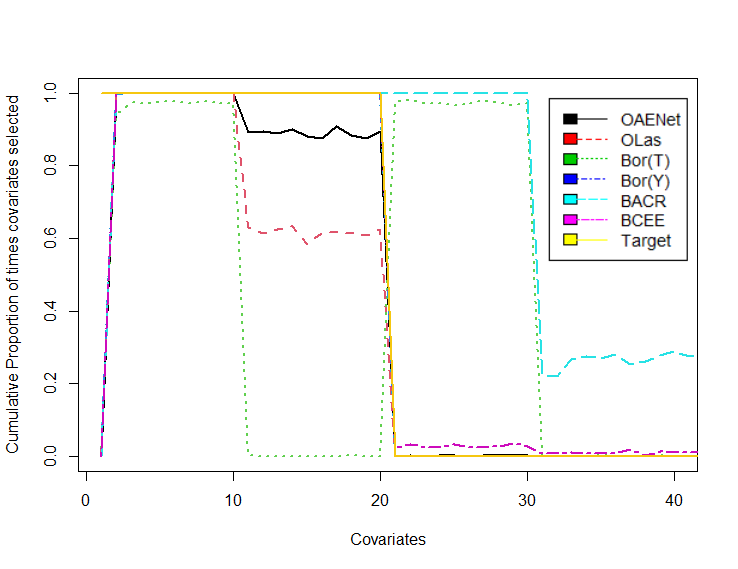}
    \caption{}
    \end{subfigure}
    \begin{subfigure}[b]{0.45\textwidth}
    \centering
    \includegraphics[width=1\textwidth]{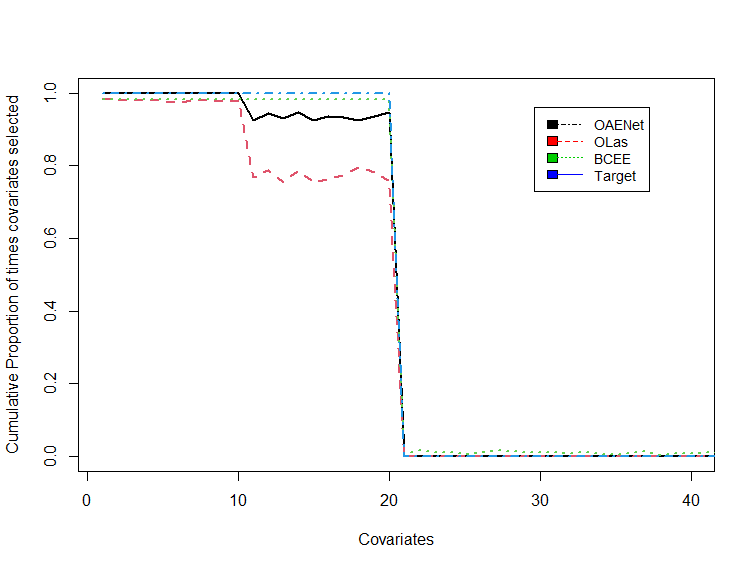}
    \caption{}
    \end{subfigure}  
    \begin{subfigure}[b]{0.45\textwidth}
    \centering
    \includegraphics[width=1\textwidth]{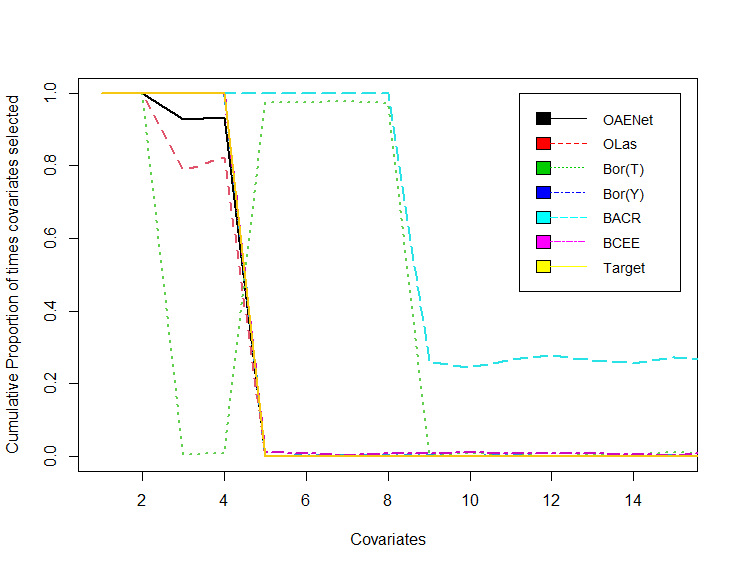}
    \caption{}
    \end{subfigure}
    \begin{subfigure}[b]{0.45\textwidth}
    \centering
    \includegraphics[width=1\textwidth]{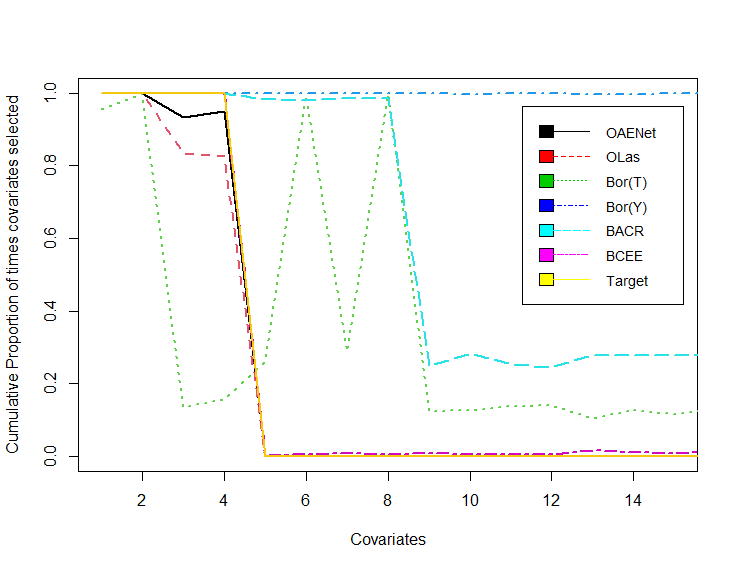}
    \caption{}
    \end{subfigure}

    \caption{Proportion of the time variables selected in (a) Scenario 1A (b) Scenario 1B (c) Scenario 2A (d) Scenario 2B.}
    \label{fig:prop}
\end{figure}

In Figure \ref{fig:prop}, we present the proportion of the time a variable is selected by the variable selection techniques for different scenarios. In scenarios 1A and 1B, our target is to select the confounders (variables $X_1,\cdots,X_{10}$) and the outcome predictors (variables $X_{11},\cdots,X_{20}$). In scenarios 2A and 2B, an ideal variable selection algorithm should select variables $X_1,\cdots,X_4$ among which, the first two are confounders and the last two are outcome predictors. In almost all cases, the outcome adaptive elastic net selects the confounders and in majority of the cases, it selects the outcome predictors. \texttt{BCEE} closely follows the target by selecting all the confounders and outcome predictors. However, it often selects some spurious variables in the process. Outcome adaptive lasso also performs well in selecting the confounders but it overly penalizes the outcome predictors. \texttt{BACR}, \texttt{Bor(T)}, and \texttt{Bor(Y)} perform poorly in this metric and often selects spurious variables.

\textcolor{black}{Another major challenge in variable selection in causal inference is to exclude weak confounders as they do not reduce bias, act as noise and inflate the variance of the estimate. Weak confounders are the variables that are strongly related to treatment but weakly associated with outcome. To compare the performance of the variable selection techniques, we added two weak confounders in Scenario 2B and ran 1000 iterations. The new outcome model is, $y = TE.A + 2X_1+2X_2+5X_3+5X_4+0.2X_7+0.2X_8$. Here, variables $X_7$ and $X_8$ are weak confounders as they are weakly associated with outcome but strongly associated with treatment. \texttt{BACR}, \texttt{Bor(T)}, and \texttt{Bor(Y)} fail to exclude the weak confounders all the time whereas \texttt{BCEE} performs slightly better than these methods but still selects the weak confounders 90\% of the times. However, \texttt{OAENet} and \texttt{OLas} do not select the weak confounders in 24\% and 1\% of the iterations, respectively. While \texttt{OLas} outperforms the \texttt{OAENet} in identifying weak confounders, in Figure 3, we see that it often over penalize and does not select some strong outcome predictors.}

\textcolor{black}{An experimental analysis of simulated data reveals that different variable selection techniques exhibit varying performance in terms of estimation quality  (bias and variance), CPU time, and selection of variables. The findings indicate that BCEE yields the highest quality estimate for the average treatment effect on the treated (ATT), but it is associated with significantly longer computation time and inconsistent adherence to theoretical expectations regarding variable selection.
In contrast, OAENet achieves an ATT estimate that closely resembles BCEE but with exceptional speed and a consistent selection of variables that align with theoretical expectations. Comparatively, other alternative methods perform poorly on one or more of the evaluated criteria. Therefore, employing OAENet is highly advantageous and potentially the optimal approach for analyzing large-scale correlated observational data. It provides a favorable trade-off between estimation quality, computational efficiency, and the selection of variables. }

\section{Case Study: Opioid Use Disorder and Suicidal Behaviour} \label{case}

In this section, we evaluate the causal relation between Opioid Use Disorder (OUD) and Suicidal behavior and present the efficiency of the proposed feature selection approach in making causal conclusions. To that end, we use the cross-sectional data from the National Survey on Drug Use and Health (NSDUH) for the years 2015 to 2019. 
NSDUH conducts yearly surveys in all 50 states, including the District of Columbia, asking people about their physical and mental health conditions, and substance use history. NSDUH data collection is sponsored by the Substance Abuse and Mental Health Services Administration (SAMHSA) with an aim to get a nationally representative estimate of mental health and drug use conditions among the USA population (age 12 and older). For the survey, household addresses are randomly selected through scientific methods to represent the population of the United States. A detailed description of the sampling strategy, data collection methods, and the questionnaire is provided in \citet{nsduh}. 

\subsection{Data Description} \label{s:methods.3}
To gauge the substance use history and health condition, participants were asked about their use and misuse of drugs, the status of mental health and substance use treatment, suicidal thoughts, and overall health condition in the past 12 months. Participants were initially asked screening questions about their history of opioid use to determine if they used prescription pain relievers at any point in their life with or without a doctor's prescription. Participants who reported misuse are categorized as ``Prescription opioid misuse'' and the rest were classified as ``Opioid use without misuse.'' If the participant agreed to opioid misuse, additional questions were asked to understand the OUD pattern, such as the frequency of use and the reason for misuse. To further understand OUD, participants were asked a set of structured questions based on the criteria of opioid abuse and dependence from the Diagnostic and Statistical Manual of Mental Disorders, Fifth Edition \citep{APA2013}. As the purpose of this study is to estimate the effect of OUD on suicidal behavior, we used recorded pain reliever abuse or dependence as the dichotomous treatment. 
To understand their thoughts over suicide, NSDUH asked participants if they ever thought about killing themselves, and did they try to kill themselves in the past 12 months. We categorized any positive response to suicidal thoughts, plans, or attempts as suicidal behavior and used it as the binary outcome.

To understand the history of non-opioid substance use of participants, questions like the age of first use, misuse frequency of drugs in the past month, misuse frequency in the past year, and recency of use of drugs. New recorded fields were created from these responses about the use of drugs like Cocaine, Crack, Heroin, Hallucinogens, Methamphetamine, Tranquilizers, Sedatives, and other stimulants in the past 12 months. NSDUH also collected data on past participation in any drug or alcohol treatment program. Participants are asked if they ever received counseling for alcohol or drugs, not counting cigarettes. Participants are also asked if they ever received mental health treatment, counseling, or support from inpatient or outpatient treatment or prescription medications for mental health treatment.

In addition, to measure the participants' mental health, NSDUH asked if they had any major depressive episodes in the past year and if they felt more depressed, anxious, or emotionally stressed. NSDUH also collected socioeconomic information of the participants, which included their age, gender, race, degree of education, employment status, income category, health insurance coverage, and population density of the location of stay. A summary of the Socioeconomic and health characteristics of participants is provided in Table \ref{nsduh}. 

\begin{sidewaystable}
\small
\caption{Socioeconomic and health characteristics of participants in NSDUH Data. The number in parenthesis represents the percentage of the sample.}\label{nsduh}
\begin{adjustbox}{scale=0.8,center}
\begin{tabular*}{\textheight}{@{\extracolsep\fill}llllll}
\toprule%
        \textbf{Covariates} & \textbf{All Participants} & \textbf{No Suicide} & \textbf{Suicide} & \textbf{OUD}  & \textbf{No OUD} \\
\midrule
 \textbf{Age} & ~ & ~ & ~ & ~ & ~ \\ \hline
        12-17 & 68089 (24.11) & 68089 (25.29) & 0 & 295 (14.58) & 67794 (24.18) \\ 
        18-25 & 69826 (24.72) & 62637 (23.26) & 7189 (54.61) & 621 (30.71) & 69205 (24.68) \\ 
        26-34 & 43976 (15.57) & 41440 (15.39) & 2536 (19.26) & 479 (23.68) & 43497 (15.51) \\ 
        35-49 & 56516 (20.01) & 54206 (20.13) & 2310 (17.54) & 446 (22.05) & 56070 (19.99) \\ 
        50+ & 43976 (15.57) & 42847 (15.91) & 1129 (8.57) & 181 (8.95) & 43795 (15.62) \\ \hline
        \textbf{Gender}  & ~ & ~ & ~ & ~ & ~ \\ \hline
        Male  & 134463 (47.61) & 128966 (47.90) & 5497 (41.75) & 1020 (50.44) & 133443 (47.59) \\ 
        Female  & 147920 (52.38) & 140253 (52.09) & 7667 (58.24) & 1002 (49.55) & 146918 (52.4) \\ \hline
        \textbf{Race}  & ~ & ~ & ~ & ~ & ~ \\ \hline
        White  & 164755 (58.34) & 156491 (58.12) & 8264 (62.77) & 1343 (66.41) & 163412 (58.28) \\ 
        African American  & 36024 (12.75) & 34643 (12.86) & 1381 (10.49) & 180 (8.9) & 35844 (12.78) \\ 
        Native American or \\ Alaskan Native & 4109 (1.45) & 3869 (1.43) & 240 (1.82) & 54 (2.67) & 4055 (1.44) \\ 
        Native Hawaiian or \\Pacific Islander & 1431 (0.5) & 1375 (0.51) & 56 (0.42) & 13 (.64) & 1418 (0.5) \\ 
        Asian  & 12959 (4.58) & 12527 (4.65) & 432 (3.28) & 34 (1.68) & 12925 (4.61) \\ 
        More than one race  & 10827 (3.83) & 10065 (3.73) & 762 (5.78) & 115 (5.68) & 10712 (3.82) \\ 
        Hispanic  & 52278 (18.51) & 50249(18.66) & 2029(15.41) & 283 (13.99) & 51995 (18.54) \\ \hline
        \textbf{Income} & ~ & ~ & ~ & ~ & ~ \\ \hline
        Less than \$20,000 & 53741 (19.03) & 49926(18.54) & 3815 (28.98) & 579 (28.63) & 53162 (18.96) \\ 
        \$20,000 - \$49,999 & 86108 (30.49) & 81722(30.35) & 4386 (33.31) & 708 (35.01) & 85400 (30.46) \\ 
        \$50,000 - \$74,999 & 43775 (15.50) & 41966 (15.58) & 1809 (13.74) & 301 (14.88) & 43474 (15.50) \\ 
        \$75,000 or more  & 98759 (34.97) & 95605 (35.51) & 3154 (23.95) & 434 (21.46) & 98325 (35.07) \\ \hline
        \textbf{Health Insurance}   & ~ & ~ & ~ & ~ & ~ \\ \hline
        Has health Insurance & 252852 (89.53) & 241575 (89.73) & 11277 (85.66) & 1698 (83.97) & 251154 (89.58) \\ \
        No health insurance & 27264(18.48) & 25467 (9.45) & 1779 (13.51) & 309 (15.28) & 26937 (9.6) \\ 
        Unavailable  & 2285 (7.73) & 2177 (0.8) & 108 (0.82) & 15 (0.74) & 2270 (0.8) \\ \hline
        \textbf{Overall health}  & ~ & ~ & ~ & ~ & ~ \\ \hline
        Excellent  & 72055 (25.51) & 70313 (26.11) & 1742 (13.23) & 181 (8.95) & 71874 (25.63) \\ 
        Very Good  & 108499 (38.42) & 104194 (38.7) & 4305 (32.70) & 630 (31.15) & 107869 (38.47) \\ 
        Good  & 74478 (26.37) & 70035 (26.01) & 4443 (33.75) & 707 (34.96) & 73771 (26.31) \\ 
        Fair/Poor & 27351 (9.68) & 24677 (9.16) & 2674 (20.31) & 504 (24.92) & 26847 (9.57) \\ \hline

\end{tabular*}
\end{adjustbox}
\end{sidewaystable}

\subsection{Evaluating Relation between OUD and Suicidal Behaviour} \label{s:methods.4}
Our dataset consists of 282,383 samples from the NSDUH data between 2015 and 2019 and includes 54 covariates. Among the analytical samples, only 2022 had OUD. The covariate set consists of information about the participant's socioeconomic, health, education, income, mental health, substance abuse history, and purpose of substance abuse. We removed covariates that had more than 25\% or more missing values. In addition, we removed the samples that had missing values in any covariate.

Furthermore, the categorical variables like education, income, and overall health were converted to binary variables using one-hot encoding. After pre-processing, the final analytical samples, after one-hot encoding, had 70 covariates and consisted of 1659 participants with recorded opioid abuse or dependency, and 208,819 participants with no opioid abuse or dependency. As the sample size for the control group is significantly higher than the treatment group, we conducted 1000 bootstrap iterations by taking 5000 random samples from the no-OUD patients (control group) in each iteration. We used Outcome Adaptive Elastic Net (\texttt{OAENet}), Outcome adaptive lasso (\texttt{OLas}), Treatment predictors (\texttt{Bor(T)}) and Outcome predictors (\texttt{Bor(Y)}) with Random Forest algorithm to select covariates. We did not use \texttt{BACR} and \texttt{BCEE} \textcolor{black}{ as BACR cannot handle categorical data due to its design limitation and
BCEE could not complete variable selection of a single iteration within couple of hours.} 
We applied Propensity Score Matching (PSM) with the Nearest Neighbor method to find one-to-one matches and calculate the Average Treatment Effect on the Treated (ATT) using the matched pairs.

In the bootstrap iterations, \texttt{ONet} selected on average 50 features. In comparison, both \texttt{Bor(T)} and \texttt{Bor(Y)} performed similarly and selected approximately 30 features on average. To analyze the performance of the variable selection algorithms, we compare the covariates selected in at least 70\% of the iterations by each algorithm to variables identified in the state-of-the-art literature. \citet{maloney2007suicidal} and \citet{streck2022national} reported that socio-economic covariates like sex, employment, and income status notably drive opioid-related mortalities, especially amongst lower-income groups. Proposed \texttt{OAENet} consistently selected these covariates, whereas \texttt{Bor(T)} and \texttt{Bor(Y)} failed to identify them. \citet{ali2021suicidal} show that race is a significant influencer of suicidal behavior. Only \texttt{OAENet} conforms to the result presented by \citet{ali2021suicidal}. According to \citet{kegler2017trends}, less urbanized areas have more suicides; however, all the algorithms did not select population density as a confounder. \texttt{OAENet} performed similar to \texttt{Bor(T)} while choosing recent pain reliever misuse frequency, which was not selected by \texttt{Bor(Y)}. Recent literature \citep{ashrafioun2017frequency,obrien2021adverse,bohnert2019understanding} show that most patient with an OUD also has a history of abusing multiple illegal substances, which is a significant risk factor of suicidal behavior. Proposed \texttt{OAENet} consistently performed better than the other algorithms in selecting covariates corresponding to illegal substance use like cocaine, crack, and heroin. Surprisingly, no algorithm selected the use of sedatives even though \citet{ashrafioun2017frequency} found it significantly associated with suicidal behavior. On the other hand, \texttt{OAENet} and \texttt{Bor(Y)} performed similarly in selecting covariates related to current mental health status, as reported by \citet{ali2021suicidal} and \citet{ashrafioun2017frequency}, whereas \texttt{Bor(T)} does not consider current mental health condition. In contrast to \texttt{OAENet}, \texttt{Bor(Y)}, and \texttt{Bor(T)}, \texttt{OLas} seems to be heavily over-penalizing and selected only five features on average. It did not select any socioeconomic variables or history of other substance abuse as a potential confounder, which contradicts existing literature. Refer to the \textbf{Online Supplement S1 Appendix A} for more information about the performance of the proposed OAENet compared to the state-of-the-art literature.

The treatment effect, ATT, an estimate for the effect of OUD on suicidal behavior with \texttt{OAENet} is 0.016 with a 95\% CI of (0.014, 0.018). ATT estimate with \texttt{OLas} is 0.032 with a 95\% CI of (0.030, 0.034), and when \texttt{Bor(Y)} was used, ATT estimate is 0.003 with a 95\% CI of (0.001, 0.006). Covariates selected with \texttt{Bor(T)} produced an ATT estimate of 0.016 with a 95\% CI of (0.014, 0.018). \textcolor{black}{\texttt{OAENet} is also almost 3 times faster to converge when compared to \texttt{Bor(T)} and \texttt{Bor(Y)} (for more information refer to \textbf{Online Supplement S1 Appendix A}).} As the estimate for the causal effect of OUD on suicidal behavior is positive and the CI does not include zero for all the methods, OUD causes suicidal behavior. Nonetheless, the small effect does not provide clinically meaningful insight.

Our effort to identify the causal relationship between OUD and suicidal behavior has several limitations. First, we used the NSDUH survey data, which utilizes a self-reporting system to collect data and, therefore, can be over- or under-reported due to the sensitive nature of the questions. Second, it is highly dependent on the participants' memory of last year's activities. Finally, we are using cross-sectional data to answer a causal question here which has many pitfalls. Hence, we cannot make any clinical or policy conclusion from this study. 

\section{Conclusion} \label{high_dim_conc}
In this paper, we discuss the variable selection problem in causal inference. Variable selection from high-dimensional data is extensively studied in machine learning literature, however, the objective of this problem in the context of causal inference is quite different. To infer causality from observational data, we have to select confounding variables which are the variables that are associated with both treatment and outcome. To that end, we propose an Outcome Adaptive Elastic Net (OAENet) method that uses a variable penalty function instead of a constant penalty function of elastic-net. First, we estimate the strength of the association of the variables to the outcome by ordinary least square regression. Then, we use the inverse of these estimates in the elastic-net regularization to create a variable penalty function. This variable regularization ensures that the outcome predictors are less penalized in the treatment model and selects all the confounders and outcome predictors. We show that the proposed OAENet method achieves the oracle property. Moreover, we evaluate the performance of the proposed technique on bias, variance, and proportion of the time right variables are selected and compare its result with state-of-the-art techniques on simulated data. Our analysis shows that the proposed outcome adaptive technique performs superiorly compared to the state-of-the-art methods such as Outcome adaptive lasso and Bayesian Regression-based methods. Finally, we present the case study with real-life policy implications. We use the NSDUH data to estimate the effect of OUD on suicidal behavior. Similar to the results on simulated data, the proposed OAENet performs better than the competing methods in covariate selection. However, the estimated causal effect of OUD on suicidal behavior is not clinically meaningful as we have several limitations in the data. In the future, we plan to conduct a longitudinal case study to identify the causal relationship between OUD and suicidal behavior. In addition, we plan to use the proposed feature selection technique for causal inference to identify high-impact societal and healthcare policy decisions.  

\section*{Acknowledgements}
Financial support from the National Science Foundation (Award Number: 2047094) is greatly acknowledged.
 \appendix
\section{Case Study - OUD and Suicidal Behaviour } \label{s:intro}
To compare the performance of the proposed Outcome Adaptive Elastic Net (OAENet) on the real-life dataset, we develop a benchmark called Expert Opinion by reviewing existing literature. First, we identify a set of journal articles \citep{ali2021suicidal, ashrafioun2017frequency, bohnert2019understanding, maloney2007suicidal, obrien2021adverse, streck2022national, moscicki1988suicide, grella2009gender, kegler2017trends} that discusses Opioid Use Disorder (OUD) and Suicidal behavior. Then, we examine the statistical models in those articles and list the variables that are found significantly associated with suicidal behavior among OUD patients. The benchmark method, Expert Opinion, is a union of all the significant variables identified in our review. Here, we are assuming that an ideal algorithm ("The Oracle") should select all the variables that are found significant in at least one of the reviewed articles. 

Our dataset considers a wide range of variables including different socio-economic, mental health, drug use/abuse related variables. On the other hand, existing literature considers a significantly smaller set of variables in their models. To present this discrepancy, we use  "NA" in the benchmark algorithm (i.e., Expert Opinion) to represent that none of the referring literature considers this covariate.

Moreover, it is important to note that, we did not conduct a systematic literature review or a meta-analysis to create the Expert Opinion as it is beyond the scope of this study. In this comparison, we plan to provide a quick evaluation of the proposed method. However, in the future, we plan to conduct a meta-analysis to create a comprehensive list of variables that contributes to suicidal behavior among OUD patients.

For the Bor(Y), OAENet, Bor(T) and OLas methods, we consider features selected in at least 70\% of the times out of the 1000 bootstrap iterations. A comparison of these methods and Expert Opinions are provided in Table \ref{tab:Summary}.  

\renewcommand{\thetable}{S\arabic{table}}

\small
\setlength{\tabcolsep}{1.9pt}
\begin{longtable}{p{6.2cm}|p{1.6cm}llll}

\caption{Comparison of performance of algorithms compared to Expert Opinion}
\label{tab:Summary} \\

\toprule
\textbf{Covariates} &\textbf{Expert Opinion} &\textbf{Bor(Y)} &\textbf{OAENet} &\textbf{Bor(T)} &\textbf{OLas}\\

\midrule
\endfirsthead
{}{\footnotesize\itshape\tablename~\thetable:
Continued from the previous page} \\
\toprule
\textbf{Covariates} &\textbf{Expert Opinion} &\textbf{Bor(Y)} &\textbf{OAENet} &\textbf{Bor(T)}&\textbf{OLas} \\
\midrule
\endhead

\midrule
{}{\footnotesize\itshape\tablename~\thetable:
Continued on next page} \\
\endfoot
\bottomrule
{\footnotesize\itshape\tablename~\thetable:
It ends from the previous page.} \\
\endlastfoot

        Ideal Case Scenario & X & X & X & X & X \\     \hline
        Age & X & X & X & ~ & ~ \\     \hline
        Gender  & X & ~ & X & ~ & ~ \\  \hline  
        Race & X & ~ & X & ~ & ~ \\     \hline
        Degree of education  & X & X & X & X & ~ \\     \hline
        Employment status & X & ~ & X & ~ & ~ \\    \hline
        Income status  & X & ~ & X & ~ & ~ \\    \hline
        Covered by health insurance  & NA & ~ & ~ & ~ & ~ \\    \hline
        Population density of where you stay  & X & ~ & ~ & ~ & ~ \\    \hline
        Overall health & X & X & X & X & ~ \\     \hline
        Most recent pain reliever misuse & X & X & ~ & X & ~  \\ \hline    
        Number of days used pain reliever not directed by the doctor in past 30 days  & NA & ~ & X & X & ~ \\    \hline
        Pain reliever misuse frequency in the last 30 days  & X & ~ & X & X & ~ \\     \hline
        Used pain reliever in greater amounts greater than past 12 months  & NA & X & X & X & ~ \\    \hline
        Used pain reliever more often than past 12 months  & NA & X & X & X & ~ \\     
        Used pain reliever longer than past 12 months  & NA & ~ & X & X & ~  \\     \hline
        Used pain reliever to relieve pain  & NA & X & X & X & ~ \\     \hline
        Used pain reliever to relax & NA & X & ~ & X & ~  \\     \hline
        Used pain reliever to experiment  & NA & ~ & X & ~ & ~  \\     \hline
        Used pain reliever to get high & NA & X & X & X & ~ \\     \hline
        Used pain reliever to sleep & NA & ~ & X & X & ~ \\     \hline
        Used pain reliever to for emotions  & NA & X & X & X & ~ \\    \hline 
        Used pain reliever to for other drug effect  & NA & ~ & X & X & ~  \\    \hline
        Used pain reliever to because hooked  & NA & X & X & X & ~  \\     \hline
        Used pain reliever to other reason  & NA & ~ & X & X & ~ \\ \hline
        Cocaine past year use  & X & X & X & ~ & ~  \\   \hline  
        Crack past year use  & X & ~ & X & ~ & ~ \\   \hline  
        Heroin past year use  & X & X & X & X & ~  \\    \hline 
        Hallucinogens past year use  & X & X & X & ~ & ~ \\     \hline
        Methamphetamine past year use  & X & X & X & X & ~ \\     \hline
        Tranquilizers past year use  & X & X & X & X & ~ \\     \hline
        Stimulants past year use  & X & X & X & ~ & ~ \\     \hline
        Sedatives past year use  & X & ~ & ~ & ~ & ~ \\     \hline
        Pain reliver ever misused  & NA & X & X & X & ~ \\     \hline
        Pain reliver misused in the past year  & NA & X & X & X & ~  \\  \hline   
        Spent more time on getting/using pain relivers in the past 12 months  & NA & X & X & X & ~ \\     \hline
        Spent lot of time on getting over the effects of prescription drugs in the past 12 months  & NA & ~ & ~ & ~ & ~ \\     \hline
        Set limits on pain reliver use in the past 12 months  & NA & X & X & X & ~ \\   \hline
        Able to keep limits you intended to keep in the past 12 months  & NA & ~ & ~ & ~ & ~ \\     \hline
        Need more pain relivers to get the same effect in the past 12 months  & NA & X & X & X & ~ \\ \hline    
        Using same amount of pain reliver had less effect in the past 12 months & NA & ~ & ~ & ~ & ~  \\    \hline 
        Want to or try cutting down the amount of pain relivers used in the past 12 months  & NA & X & X & X & ~ \\    \hline 
        Able to cut or stop using pain relivers in the past 12 months  & NA & ~ & ~ & ~ & ~ \\    
        Able to cut down or stop using pain relivers once in the past 12 months  & NA & ~ & ~ & ~ & ~ \\    \hline 
        Had 3+ pain reliver withdrawal symptoms  & NA & ~ & ~ & ~ & ~ \\     \hline
        Recorded pain reliver abuse or dependence in the past year  & X & ~ & ~ & ~ & ~ \\     \hline
        Ever received alcohol or drug treatment in the past year  & X & ~ & ~ & ~ & ~ \\    \hline
        Past year did you have any major depressive episode  & X & X & X & ~ & X   \\     \hline
        Received any mental health treatment in the past year  & X & X & X & ~  & ~ \\     \hline
        In the past 30 days felt more depressed, anxious or emotionally stressed  & X & X & X & ~ & ~  \\     \hline
        Year of the data made  & ~ & ~ & X & ~ & ~  \\   

\end{longtable}

\begin{table}[!ht]
    \centering
    \caption{Computation time required for algorithms to calculate ATT in seconds}
    \begin{tabular}{lcccc}
    \hline
        \textbf{Algorithm} & \textbf{OAENet} & \textbf{Olas} & \textbf{Bor(T)} & \textbf{Bor(Y)} \\ \hline
        \textbf{Time} & 32.02 & 0.74 & 90.59 & 221 \\ \hline
    \end{tabular}
\end{table}

\section{ Proof Theorem 3.1} \label{s:theorem}
The proof of this theorem is adopted from \citep{ghosh2011grouped}. 

Let's assume $C_n = \frac{1}{n}X^TX \rightarrow C$, where $C$ is a positive definite matrix and $|A| = p_0 < P$ where $P$ is the total number of variables in the covariate set. Without loss of generality, we can say that $A = \left \{ 1,2, \cdots, p_0 \right \}$.
Let $C =  \begin{pmatrix}
  C_{11} & C_{12}\\ 
  C_{21} & C_{22}
\end{pmatrix}$
where $C_{11}$ is a $P_0 \times P_0$ matrix. $C_{11}$ is also known as the Fisher Information Matrix.

Let's define $A^* = \left\{ j, \hat{\beta_j}^* \neq 0   \right\} $, now we define the maximum likelihood equation for the Outcome Adaptive Lasso: 

\begin{align}
    \hat{\boldsymbol{\beta}}^*(OLas) = \left\{ {\argmin}_{\boldsymbol\beta} \left \Vert\mathbf{y}^*-\mathbf{X^*\beta}  \right\Vert^2_2 + \lambda_1 \sum_{j=1}^p \hat{w}_j\left \Vert\boldsymbol{\beta}\right\Vert_1^1 \right\} \label{oeanet} ,\\
    {{\boldsymbol{X}^*}_{(n+p)\times p} = \begin{pmatrix}
    \boldsymbol{X} \\ 
     \sqrt{\lambda_2}\boldsymbol{I}
    \end{pmatrix}} ,
    {\boldsymbol{y^*}_{(n+p)\times 1} = \begin{pmatrix}
  \boldsymbol{y} \\
  \boldsymbol{0}
\end{pmatrix}}
\end{align}

On the artificial data set $(\boldsymbol{y^*}, \boldsymbol{X^*})$ we make the following assumptions: 
\begin{enumerate}
  \item Given the original dataset $(\boldsymbol{y}, \boldsymbol{X})$ and fixed $(\lambda_1, \lambda_2)$ we would like to define an artificially augmented data set as $(\boldsymbol{y^*}, \boldsymbol{X^*})$
  \item $C^*$ as $C^*_n = \frac{1}{n}X^{*T}X^* = C_n + \frac{\lambda_2^{(n)}}{n}$ $\rightarrow C$, where $C$ is a positive definite matrix and $\lambda_2/n \rightarrow 0$ asymptotically.
\end{enumerate}

From the outcome adaptive lasso formulation of outcome adaptive elastic net it shows that \eqref{oeanet} is still a convex optimization problem and has the same optimization properties as outcome adaptive lasso. 

In case of orthogonal design it is straightforward to show that with parameter $(\lambda_1,\lambda_2, \boldsymbol{\hat{W}})$ the solution of \eqref{oeanet} is given by:
\begin{align}
    \hat{\beta^*}(OLas) = (1+\lambda_2)^{-1}(\Vert\hat{\beta}(ols)\Vert-\hat{w}\lambda_1/2)_+sgn \{\hat{\beta(ols)}\} \label{eq3}
\end{align}

where $\boldsymbol{\hat{\beta}}(ols) = \boldsymbol{X^Ty}$ and $z_+$ denotes the positive part of z; $z_+=z$ if $z>0$ and 0 otherwise. Moving forward to rescale the estimate \eqref{eq3} to give us outcome adaptive elastic net as:

\begin{align}
    \hat{\beta}(OAENet) = (1+\lambda_2)\hat{\beta^*}(OLas)
\end{align}

It is evident from \eqref{eq3} that solution to outcome adaptive elastic net estimate is a rescaled version of outcome adaptive lasso. 

\citet{shortreed2017outcome} shows that the Outcome-adaptive LASSO problem can achieve the oracle properties. Therefore, we can claim that the proposed Outcome Adaptive Elastic Net (OAENet) also achieves the oracle property by satisfying the consistency in variable selection and asymptotic normality. 

\bibliographystyle{elsarticle-harv} 
\bibliography{bib}


\end{document}